\documentclass[sigconf]{acmart}
\AtBeginDocument{%
  }

\copyrightyear{2023}
\acmYear{2023}
\setcopyright{acmlicensed}\acmConference[DocIU Workshop, CIKM '23]{Proceedings of the 32nd ACM International Conference on Information and Knowledge Management}{October 21--25, 2023}{Birmingham, United Kingdom} \acmBooktitle{Proceedings of the 32nd ACM International Conference on Information and Knowledge Management (DocIU Workshop, CIKM '23), October 21--25, 2023, Birmingham, United Kingdom}
\acmDOI{https://doi.org/XXXXXX.XXXX}

\begin{document}


\title{Jaeger: A Concatenation-Based Multi-Transformer VQA Model}
\author{Jieting Long}
\authornote{All four authors contributed equally to this research.}
\affiliation{%
  \institution{The University of Sydney}
  \city{Sydney}
  \state{NSW}
  \country{Australia}
  \postcode{2050}
}
\email{jieting.long@sydney.edu.au}

\author{Zewei Shi}
\authornotemark[1]
\affiliation{%
  \institution{The University of Sydney}
  \city{Sydney}
  \state{NSW}
  \country{Australia}
  \postcode{2050}
}
\email{zewei.shi@sydney.edu.au}

\author{Penghao Jiang}
\authornotemark[1]
\affiliation{%
  \institution{The University of Sydney}
  \city{Sydney}
  \state{NSW}
  \country{Australia}
  \postcode{2050}
}
\email{pjia0498@uni.sydney.edu.au}

\author{Yidong Gan}
\authornotemark[1]
\affiliation{%
  \institution{The University of Sydney}
  \city{Sydney}
  \state{NSW}
  \country{Australia}
  \postcode{2050}
}
\email{yidong.gan@sydney.edu.au}

\renewcommand{\shortauthors}{Long et al.}


\begin{abstract}
Document-based Visual Question Answering poses a challenging task between linguistic sense disambiguation and fine-grained multimodal retrieval. Although there has been encouraging progress in document-based question answering due to the utilization of large language and open-world prior models\cite{1}, several challenges persist, including prolonged response times, extended inference durations, and imprecision in matching. In order to overcome these challenges, we propose Jaegar, a concatenation-based multi-transformer VQA model. To derive question features, we leverage the exceptional capabilities of RoBERTa large\cite{2} and GPT2-xl\cite{3} as feature extractors. Subsequently, we concatenate the outputs from both models. This operation allows the model to consider information from diverse sources concurrently, strengthening its representational capability. By leveraging multiple pre-trained models for feature extraction, our approach has the potential to amplify their performance through concatenation. After concatenation, we apply dimensionality reduction to the output features, reducing the model's computational overhead and inference time. Empirical results demonstrate that our proposed model achieves competitive performance on Task C of the PDF-VQA Dataset. If the user adds any new data, they should make sure to style it as per the instructions provided in previous sections. 
\end{abstract}



\maketitle

\section{Introduction}
Document-based Visual Question Answering has various applications in the fields of biomedical science, business, and education. However, addressing the accuracy of answering questions and the inference time has always been crucial. To address this challenge, we leverage large language models' robust feature extraction capabilities to identify the correct answers to questions precisely. We aim to develop a Document-based Visual Question-answering model that relies on large language models to minimize inference time and reduce computational resource demands.
Previous studies have investigated different approaches to Document-based Visual Question Answering, most of which utilized CNN and LSTM models to specifically address questions about image content\cite{4}. However, these models faced difficulties in handling complex relationships and contextual information, resulting in below-average performance on intricate questions. As attention mechanism techniques have advanced, more studies have chosen to incorporate attention mechanisms into models to improve their interpretability and performance. This allows a model to automatically focus on question-relevant areas, enhancing its ability to handle complex queries\cite{5}. However, this also significantly increases the complexity and computational requirements of the model. Subsequently, the emergence of multi-modal models introduced a new research direction. Multi-modal models can capture intricate relationships between images and text, often outperforming single-modal models\cite{6}. Nonetheless, most multi-modal models require substantial amount of training data to avoid overfitting. When abundant data is unavailable, multi-modal models may have limited applicability to general downstream tasks.
We summarize our contributions as follows:
\begin{itemize}
    \item We introduce Jaeger, a novel model for document-based visual question answering model.
    \item We design a concatenation strategy to enhance the model’s performance and presentation capabilities.
    \item We demonstrate that Jaeger can output competitive results on multiple tasks.
\end{itemize}

\section{Related Work}
Large language models are pretrained on vast amounts of text data, exhibiting comprehensive understanding of natural language. Existing studies have introduced advanced training techniques to optimize their ability to generalize across diverse tasks. Johnson et al. propose adaptive training techniques for large language models and various architecture improvements. They further leverage the potential of transfer learning, showing significant results in downstream tasks like text classification and sentiment analysis. These recent advancements pave the way for the extensive application of large language models in various domains, including machine translation\cite{7}, chatbots\cite{8}, text summarization\cite{9} , code generation\cite{10}, medical diagnosis from textual data, and more. The ubiquity and versatility of such models underline their importance in the current AI landscape.

\section{Methodology}
\subsection{Problem Definition}
Hierarchical relationship understanding stands as one of the key challenges in PDF VQA tasks. The primary objective is to augment comprehension at the document level, focusing on two question types: understanding parent/child relationships. That is, all contents hierarchically related to the queried item in the question are expected to be identified\cite{11}.

\subsection{Framework}

\begin{figure}[h]
  \centering
  \includegraphics[width=\linewidth]{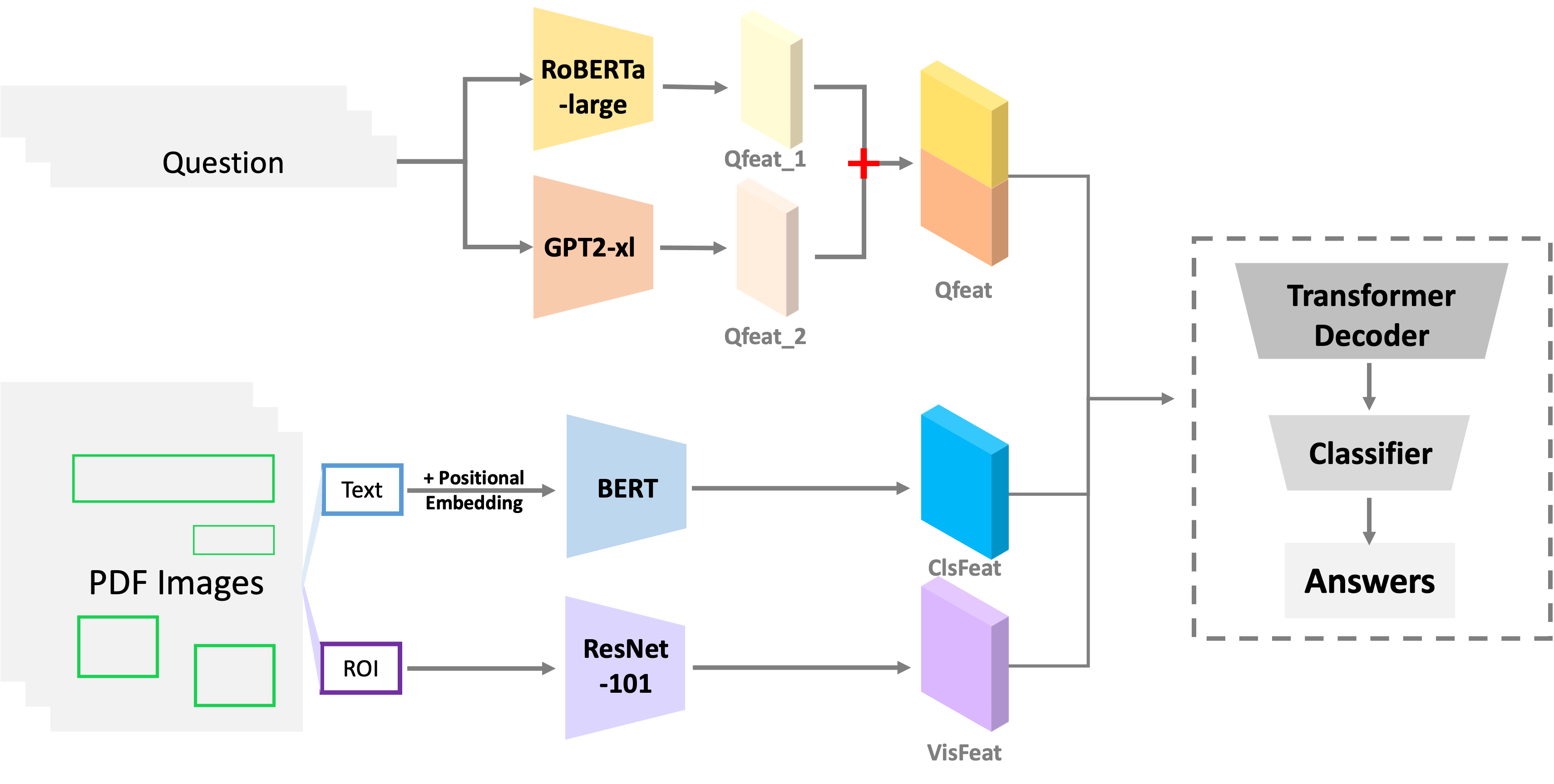}
  \label{arch}
  \caption{Model Architecture. All encoders used are pretrained models (i.e., pretrained RoBERTa-large, pretrained GPT2-xl, pretrained BERT, and pretrained ResNet-101). Qfeat is the result of concatenating Qfeat\_1 and Qfeat\_2 which are the last hidden state from the corresponding encoder. }
\end{figure}

Our framework, Jaeger, is dedicated to acquiring distinct and valuable representations through the use of pretrained models. It encompasses the extraction of both textual and visual features, as shown in Figure 1. Textual features are obtained from two perspectives: (1) questions and (2) the primary content, while visual features are derived from region-of-interest areas within each page. Regarding questions, we leverage two pretrained large language models (i.e., RoBERTa-large and GPT2-xl) to extract features, concatenating them to capture information from different aspects effectively. For PDFs, we access content text incorporating positional information after tokenization and region-of-interest identification. We then utilize pretrained models, specifically Bert and ResNet-101, to extract textual and visual features, respectively. This results in three distinct sets of features, each serving as a representation for a specific aspect of the problem.

\section{Experiments}
\subsection{Dataset and Experimental Settings}
We employ a publicly available PDF VQA dataset\cite{11}, which includes a collection of question-answers pairs with specified question types, textual content, positional information, and each PDF page is represented as an image. Our chosen evaluation metric is Exact Matching Accuracy (EMA), which considers a prediction as accurate if only if it matches all ground truth answers for a given question. We utilize bert-base-uncased to tokenize and employ the SGD optimizer, setting the learning rate at 1e-06.
\subsection{Baseline and Performance Comparison}
We compare our model with three large visual-and-language pretrained models (VLPMs) - VisualBERT, ViLT, and LXMERT, alongside the state-of-the-art method (LoSpa) on the PDF-VQA challenge. The distinguishing factor of Jaeger lies in its question feature processing step, as opposed to all other methods, which process questions as sequences of question tokens encoded by pretrained BERT models\cite{11}. As depicted in Table 1, our Jaeger model demonstrates the highest performance when evaluated on both the validation and testing sets, outperforming all baseline models.

\begin{table}[h]
  \begin{tabular}{ccc}
    \toprule
    Model & Val & Test\\
    \midrule
    VisualBERT\cite{12} & 21.55& 18.52 \\
    ViLT\cite{13}& 10.21 & 9.87\\
    LXMERT\cite{14}& 16.37& 14.41\\
    LoSpa\cite{11}& 30.21& 28.99\\
    \midrule
    Our Jaeger & 35.87 & 33.63\\
    \bottomrule
  \end{tabular}
  \vspace{2mm}
  \caption{Performance Comparison Using the EMA Metric}
  \label{result}
\end{table}
\vspace{-5mm}
\section{Conclusion and Future Work}
In this paper, we have introduced Jaeger, a novel model for document based VQA tasks that leverages the robust feature extraction capabilities of large language models. Through a carefully designed concatenation strategy, the model achieves a new state-of-the-art. Our work underscores the potential of combining state-of-the-art language models to address the challenges of document based VQA.
As for future work, we will explore relational features extraction methods. We believe enhancing the model’s structural understanding at the document level will facilitate the identification of hierarchically related items, leading to a higher EMA. Additionally, we will also examine the number of large language models to be involved when extracting features. Furthermore, we plan to investigate new fine-tuning strategies tailored to our specific task, aiming to further boost performance while maintaining efficiency would also be valuable.

\bibliographystyle{ACM-Reference-Format}
\bibliography{jaeger_usydAYN}


\end{document}